\documentclass[11pt,a4paper]{article}
\usepackage[hyperref]{emnlp-ijcnlp-2019}
\usepackage{times}
\usepackage{latexsym}

\usepackage{url}
\usepackage{helvet}  
\usepackage{courier}  
\usepackage{url}  
\usepackage{graphicx}  
\usepackage{natbib}
\usepackage{color}
\usepackage{amsfonts}
\usepackage{amsmath}
\usepackage{multirow}
\usepackage{booktabs}
\usepackage{comment}
\usepackage{caption}
\usepackage{subcaption}

\aclfinalcopy 

\title{Multilingual Neural Machine Translation with Language Clustering}

\author{Xu Tan$^\S$\thanks{Authors contribute equally to this work.}, Jiale Chen$^\dag$\footnotemark[1], Di He$^\ddag$, Yingce Xia$^\S$, Tao Qin$^\S$ \and Tie-Yan Liu$^\S$ \\
$^\S$Microsoft Research \\
$^\dag$University of Science and Technology of China \\
$^\ddag$Peking University \\
\{xuta,taoqin,tyliu\}@microsoft.com  \\
}

\date{}

\begin{document}
\maketitle
\begin{abstract}
Multilingual neural machine translation (NMT), which translates multiple languages using a single model, is of great practical importance due to its advantages in simplifying the training process, reducing online maintenance costs, and enhancing low-resource and zero-shot translation.  Given there are thousands of languages in the world and some of them are very different, it is extremely burdensome to handle them all in a single model or use a separate model for each language pair. Therefore, given a fixed resource budget, e.g., the number of models, how to determine which languages should be supported by one model is critical to multilingual NMT, which, unfortunately, has been ignored by previous work. In this work, we develop a framework that clusters languages into different groups and trains one multilingual model for each cluster. We study two methods for language clustering: (1) using prior knowledge, where we cluster languages according to language family, and (2) using language embedding, in which we represent each language by an embedding vector and cluster them in the embedding space. In particular, we obtain the embedding vectors of all the languages by training a universal neural machine translation model. Our experiments on 23 languages show that the first clustering method is simple and easy to understand but leading to suboptimal translation accuracy, while the second method sufficiently captures the relationship among languages well and improves the translation accuracy for almost all the languages over baseline methods.

\end{abstract}

\section{Introduction}
\noindent Neural machine translation (NMT)~\citep{DBLP:journals/corr/BahdanauCB14,DBLP:conf/emnlp/LuongPM15,DBLP:conf/nips/SutskeverVL14,DBLP:journals/corr/WuSCLNMKCGMKSJL16,DBLP:conf/icml/GehringAGYD17,DBLP:conf/nips/VaswaniSPUJGKP17} has witnessed rapid progress in recent years, from novel model structure developments~\citep{DBLP:conf/icml/GehringAGYD17,DBLP:conf/nips/VaswaniSPUJGKP17} to achieving performance comparable to humans~\citep{DBLP:journals/corr/abs-1803-05567}. 

Although a conventional NMT model can handle a single language translation pair (e.g., German$\to$English, Spanish$\to$French) well, training a separate model for each language pair is unaffordable considering there are thousands of languages in the world. A straightforward solution to reduce computational cost is using one model to handle the translations of multiple languages, i.e., multilingual translation. ~\citet{DBLP:journals/tacl/JohnsonSLKWCTVW17,DBLP:conf/naacl/FiratCB16,DBLP:journals/corr/HaNW16,DBLP:journals/corr/abs-1804-08198} propose to share part of (e.g., attention mechanism)  or all models for multiple language pairs and achieve considerable accuracy improvement. While they focus on how to translate multiple language pairs in a single model and improve the performance of the multilingual model, they do not investigate which language pairs should be trained in the same model. 


Clearly, it is extremely heavy to translate all language pairs in a single model due to the diverse and large amount of languages; instead we cluster language pairs into multiple clusters and train one model for each cluster, considering that: (1) language pairs that differ a lot (e.g., German$\to$English and Chinese$\to$English) may negatively impact the training process if handling them by one model, and (2) similar language pairs (e.g., German$\to$English and French$\to$English) are likely to boost each other in model training. Then the key challenge is how to cluster language pairs, which is our focus in this paper.

In this paper, we consider the many-to-one settings where there are multiple source languages and one target language (English), and one-to-many settings where there are multiple target languages and one source language (English)\footnote{Many-to-many translation can be bridged through many-to-one and one-to-many translations. Our methods can be also extended to the many-to-many setting with some modifications. We leave this to future work.}. In this way, we only need to consider the languages in the source or target side for the determination of training in the same model, instead of language pairs.

We consider two methods for language clustering. The first one is clustering based on prior knowledge, where we use the knowledge of language family to cluster languages. The second one is a purely learning based method that uses language embeddings for similarity measurement and clustering. We obtain the language embeddings by training all the languages in a universal NMT model, and add each language with a tag to give the universal model a sense of which language it currently processes. The tag of each language (language embedding) is learned end-to-end and used for language clustering. Language clustering based on language family is easy to obtain and understand, while the end-to-end learning method may boost multilingual NMT training, since the language clustering is derived directly from the learning task and will be more useful to the task itself.

Through empirical studies on 23 languages$\to$English and English$\to$23 languages in IWSLT dataset, we have several findings: (1) Language embeddings can capture the similarity between languages, and correlate well with the fine-grained hierarchy (e.g., language branch) in a  language family. (2) Similar languages, if clustered together, can indeed boost multilingual performance. (3) Language embeddings based clustering outperforms the language family based clustering on the 23 languages$\leftrightarrow$English.

\section{Background}
\paragraph{Neural Machine Translation} Given the bilingual translation pair $(x, y)$, an NMT model learns the parameter $\theta$ by maximizing the log-likelihood $\log P(y|x, \theta)$. The encoder-decoder framework~\citep{DBLP:journals/corr/BahdanauCB14,DBLP:conf/emnlp/LuongPM15,DBLP:conf/nips/SutskeverVL14,DBLP:journals/corr/WuSCLNMKCGMKSJL16,DBLP:conf/icml/GehringAGYD17,DBLP:conf/nips/VaswaniSPUJGKP17,shen2018dense,he2018layer} is adopted to model the conditional probability $P(y|x, \theta)$, where the encoder maps the input to a set of hidden representations $h$ and the decoder
generates each target token $y_t$ using the previous generated tokens $y_{<t}$ and the representations $h$.

\paragraph{Multilingual NMT} NMT has recently been extended from the translation of a language pair to multilingual translation~\citep{dong2015multi,DBLP:journals/corr/LuongLSVK15,DBLP:conf/naacl/FiratCB16,DBLP:journals/corr/abs-1804-08198,DBLP:journals/tacl/JohnsonSLKWCTVW17,he2019language,DBLP:journals/corr/HaNW16,platanios2018contextual,tan2018multilingual}.~\citet{dong2015multi} use a single encoder but separate decoders to translate one source language to multiple target languages (i.e., one-to-many translation).~\citet{DBLP:journals/corr/LuongLSVK15} combine multiple encoders and decoders, one encoder for each source language and one decoder for each target language respectively, to translate multiple source languages to multiple target languages (i.e., many-to-many translation).~\citet{DBLP:conf/naacl/FiratCB16} use different encoders and decoders but share the attention mechanism for many-to-many translation. Similarly,~\citet{DBLP:journals/corr/abs-1804-08198} propose the neural interlingua, which is an attentional LSTM encoder to link multiple encoders and decoders for different language pairs. ~\citet{DBLP:journals/tacl/JohnsonSLKWCTVW17,DBLP:journals/corr/HaNW16} use a universal encoder and decoder to handle multiple source and target languages, with a special tag in the encoder to determine which target languages to output. 

While all the above works focus on the design of better models for multilingual translation, they implicitly assume that a set of languages are pre-given, and do not consider which languages (language pairs) should be in a set and share one model. In this work we focus on determining which languages (language pairs) should be shared in one model.

\section{Multilingual NMT with Language Clustering}
Previous work trains a set of language pairs (usually the number is small, e.g., $<10$ languages) using a single model and focuses on improving this multilingual model. When facing a large amount of languages (dozens or even hundreds), one single model can hardly handle them all, considering that a single model has limited capacity, some languages are quite diverse and far different, and thus one model will lead to degraded accuracy. 

In this paper, we first group the languages into several clusters and then train one multilingual NMT model~\citep{DBLP:journals/tacl/JohnsonSLKWCTVW17} for the translations in each cluster. By controlling the number of clusters, we can balance translation accuracy and computational cost: compared to using one model for each language pair, our approach greatly reduces computational cost, and compared to using one model for all language pairs, our approach delivers better translation accuracy. In this section, we present two methods of language clustering to enhance multilingual NMT.

\subsection{Prior Knowledge Based Clustering} 
Several kinds of prior knowledge can be leveraged to cluster languages~\citep{paul2009ethnologue,comrie1989language,bakker2009adding,holman2008explorations,liu2010language,chen2017classifying,wu2018beyond,leng2019unsupervised}, such as language family, language typology or other language characteristics from the URIEL database~\citep{DBLP:conf/eacl/LevinLMLKT17}. Here we do not aim to give a comprehensive study of diverse prior knowledge based clusterings. Instead, we choose the commonly used language family taxonomy~\citep{paul2009ethnologue} as the prior knowledge for clustering, to provide a comparison with the language embedding based clustering. 

Language family is a group of languages related through descent from a common ancestral language or parental language\footnote{https://en.wikipedia.org/wiki/Language\_family}. There are different kinds of taxonomies for language family in the world, among which, Ethnologue~\citep{paul2009ethnologue}\footnote{https://www.ethnologue.com/browse/families} is one of the most authoritative and commonly accepted taxonomies. The 7,472 known languages in the world fall into 152 families according to~\citet{paul2009ethnologue}. We regard the languages in the same family as similar languages and group them into one cluster.

\subsection{Language Embedding Based Clustering} 

When training a multilingual model, it is common practice to add a tag to the input of encoder to indicate which language the model is currently processing~\citep{DBLP:journals/tacl/JohnsonSLKWCTVW17,DBLP:conf/naacl/FiratCB16,DBLP:conf/emnlp/MalaviyaNL17,DBLP:conf/eacl/TiedemannO17}. The embeddings of the tag is learned end-to-end and depicts the characteristics of the corresponding language, which is called language embeddings, analogous to word embeddings~\citep{mikolov2013efficient}. 

\begin{figure}
  \centering
  \includegraphics[width=0.48\textwidth]{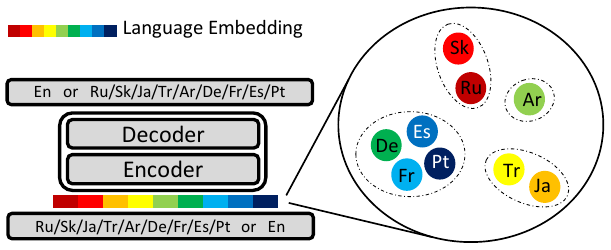}
  \caption{ The illustration of learning language embeddings for clustering. For both many-to-one (other languages to English) and one-to-many (English to other languages) setting, we add the language embeddings to the encoder.}
  \label{archi_lan_embed}
\end{figure}

We first train a universal model to translate all the language pairs. As shown in Figure~\ref{archi_lan_embed}, the encoder of the universal NMT model takes both word embeddings and language embedding as inputs. After model training, we get an embedding vector for each language. We regard the embedding vector as the representation of a language and cluster all languages using the embeddings\footnote{One may argue that considering training one universal model for all the languages can not ensure the best translation accuracy, how could the learned language embeddings be accurate? We find the learned language embeddings are not that sensitive to translation accuracy and relatively stable during the later training process even when the universal model has not converged to the best translation accuracy. Therefore, the cost to get the language embeddings is much smaller than training a multilingual model to get good accuracy.}. 
There exist many clustering methods. Without loss of generality, we choose hierarchical clustering~\citep{rocach2005clustering} to cluster the embedding vectors in our experiments.

\subsection{Discussions} 
We analyze and compare the two clustering methods proposed in this section. Clustering based on prior knowledge (language family) is simple. The taxonomy based on language family is consistent with the human knowledge, easy to understand, and does not change with respect to data/time. This method also has drawbacks. First, the taxonomy built on language family does not cover all the languages in the world since some languages are isolate\footnote{https://en.wikipedia.org/wiki/Language\_isolate}. Second, there are many language families (152 according to~\citet{paul2009ethnologue}), which means that we still need a large number of models to handle all the languages in the world. Third, language family cannot characterize all the features of a language to fully capture the similarity between languages.

Since language embeddings are learnt in a universal NMT model, which is consistent with the downstream multilingual NMT task, clustering based on language embeddings is supposed to capture the similarity between different languages well for NMT. As is shown in our experiments, the clustering results implicitly capture and combine multiple aspects of language characteristics and boost the performance of the multilingual model. 

\section{Experiment Setup}
\paragraph{Datasets} We evaluate our method on the IWSLT datasets which contain multiple languages from TED talks. We collect datasets from the IWSLT evaluation campaign\footnote{https://wit3.fbk.eu/} from years 2011 to 2018, which consist of the translation pairs of 23 languages$\leftrightarrow$English. The detailed description of the training/validation/test set of the 23 translation pairs can be found in Supplementary Materials (Section 1). All the data has been tokenized and segmented into sub-word symbols using Byte Pair Encoding (BPE)~\citep{DBLP:conf/acl/SennrichHB16a}. We learn the BPE operations for all languages together, which results in a shared vocabulary of 90K BPE tokens.

\paragraph {Model Configurations}
We use Transformer~\citep{DBLP:conf/nips/VaswaniSPUJGKP17} as the basic NMT model considering that it achieves state-of-the-art performance on multiple NMT benchmark tasks and is a popular choice for recent research on NMT. We use the same model configuration for each cluster with model hidden size $d_{\text{model}}=256$, feed-forward hidden size $d_{\text{ff}}=1024$ and the layer number is 2. The size of language embeddings is set to $256$.

\paragraph {Training and Inference}
We up-sample the training data of each language to be roughly the same during training. For the multilingual model training, we concatenate training sentence pairs of all the languages in the same mini-batch. We set the batch size of each language to roughly 4096 tokens, and thus the total batch size is $4096 * |C_k|$, where $|C_k|$ is the number of languages in cluster $k$. The corresponding language embedding is added on the word embedding of each source token. We train the model with 8 NVIDIA Tesla M40 GPU cards, each GPU card with roughly $512 * |C_k|$ tokens in terms of batch size. We use Adam optimizer~\citep{kingma2014adam} with $\beta_{1}= 0.9$, $\beta_{2} = 0.98$, $\varepsilon = 10^{-9}$ and follow the learning rate schedule in \citet{DBLP:conf/nips/VaswaniSPUJGKP17}. 

During inference, each source token is also added with the corresponding language embedding in order to give the model a sense of the language it is currently processing. We decode with beam search and set beam size to 6  and length penalty $\alpha=1.1$ for all the languages. We evaluate the translation quality by tokenized case sensitive BLEU~\citep{DBLP:conf/acl/PapineniRWZ02} with multi-bleu.pl\footnote{https://github.com/moses-smt/mosesdecoder/blob/
master/scripts/generic/multi-bleu.perl}.

Our codes are implemented based on tensor2tensor~\citep{tensor2tensor}\footnote{https://github.com/tensorflow/tensor2tensor} and we will release the codes once the paper is open to the public.

\section{Results}
In this section, we mainly show the experiment results and analyses on the many-to-one setting in Section~\ref{ret_lan_cluster}-\ref{ret_vary_data}. The results on the one-to-many setting are similar and we briefly show the results in Section~\ref{ret_one_to_many} due to space limitations.

\begin{figure}
  \centering
  \includegraphics[width=0.5\textwidth]{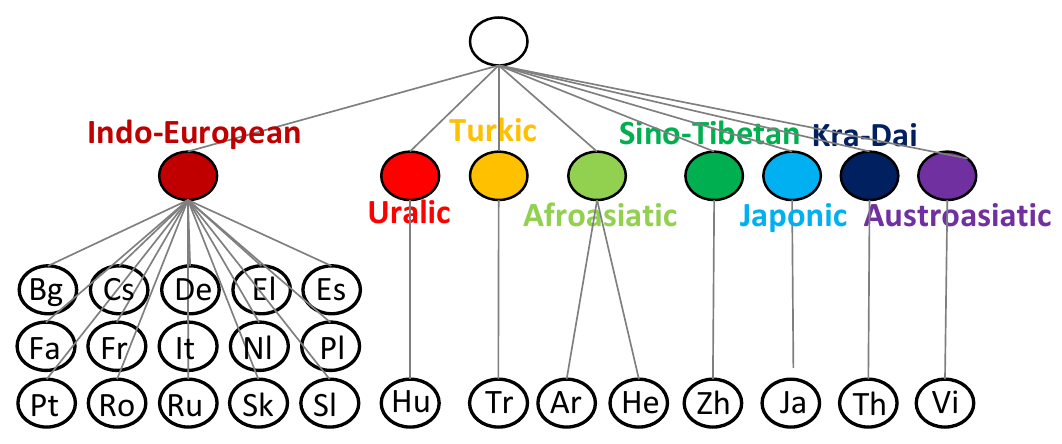}
  \caption{ Language clustering of the 23 languages in IWLST dataset according to language family. There are 8 different language families in this dataset, which includes Indo-European, Uralic, Turkic, Afroasiatic, Sino-Tibetan, Japonic, Kra-Dai and Austroasiatic.}
  \label{cluster_prior_family}
\end{figure}

\subsection{Results of Language Clustering}
\label{ret_lan_cluster}
The language clustering based on language family is shown in Figure~\ref{cluster_prior_family}, which results in 8 groups given our 23 languages. All the language names and their corresponding ISO-639-1 code\footnote{https://www.iso.org/iso-639-language-codes.html.} can be found in Supplementary Materials (Section 2). 

\begin{figure*}
  \centering
  \includegraphics[width=1.0\textwidth]{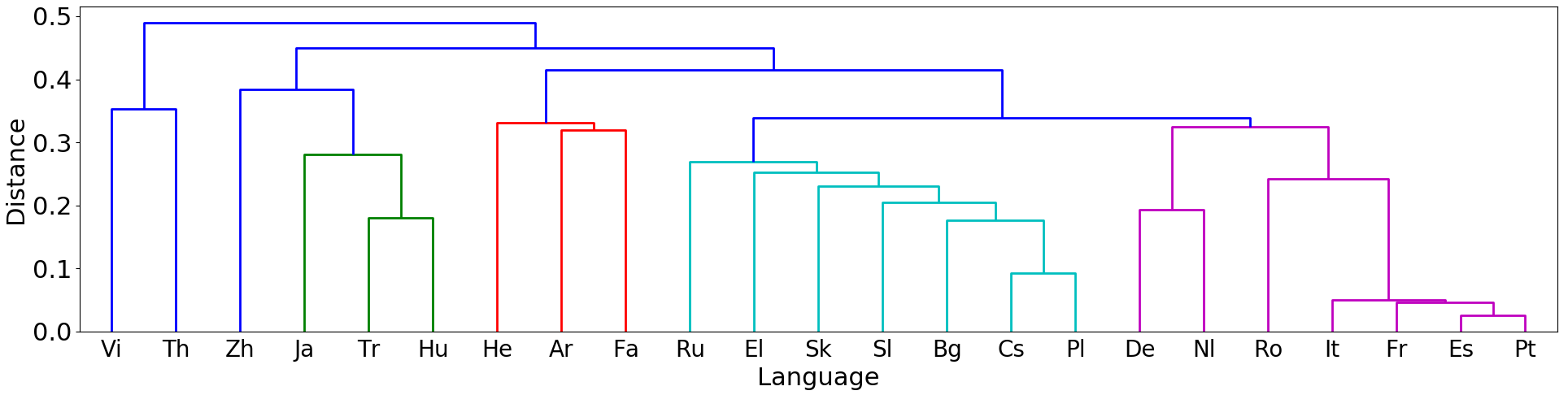}
  \caption{ The hierarchical clustering based on language embeddings. The Y-axis represents the distance between two languages or clusters. Languages in the same color are divided into the same cluster, where blue color agglomerates different clusters together. If a language is marked as blue, then it forms a cluster itself. Cluster \#1: Vi; Cluster \#2: Th; Cluster \#3: Zh; Cluster \#4: Ja, Tr, Hu; Cluster \#5: He, Ar, Fa; Cluster \#6: Ru, El, Sk, Sl, Bg, Cs, Pl; Cluster \#7: De, Nl, Ro, It, Fr, Es, Pt.}
  \label{24_NMT_hirach_average_cosine_cluster5}
\end{figure*}

We use hierarchical clustering~\citep{rocach2005clustering}\footnote{https://docs.scipy.org/doc/scipy-0.18.1/reference/generated/
scipy.cluster.hierarchy.linkage.html} method to group the languages based on language embeddings. We use the elbow method~\citep{thorndike1953belongs} to automatically decide the optimal number of clusters $K$. Note that we have tried to extract the language embeddings from multiple model checkpoints randomly chosen in the later training process, and found that the clustering results based on these language embeddings are stable. Figure~\ref{24_NMT_hirach_average_cosine_cluster5} demonstrates the clustering results based on language embeddings. Each color represents a language cluster and there are 7 clusters according to the elbow method (the details of how this method determines the optimal number of clusters are shown in Figure~\ref{determin_cluster}). Figure~\ref{24_NMT_hirach_average_cosine_cluster5} clearly shows the agglomerative process of the languages and demonstrates the fine-grained relationship between languages.

We have several interesting findings from Figure~\ref{cluster_prior_family} and~\ref{24_NMT_hirach_average_cosine_cluster5}:

\begin{figure}
  \centering
  \includegraphics[width=0.32\textwidth]{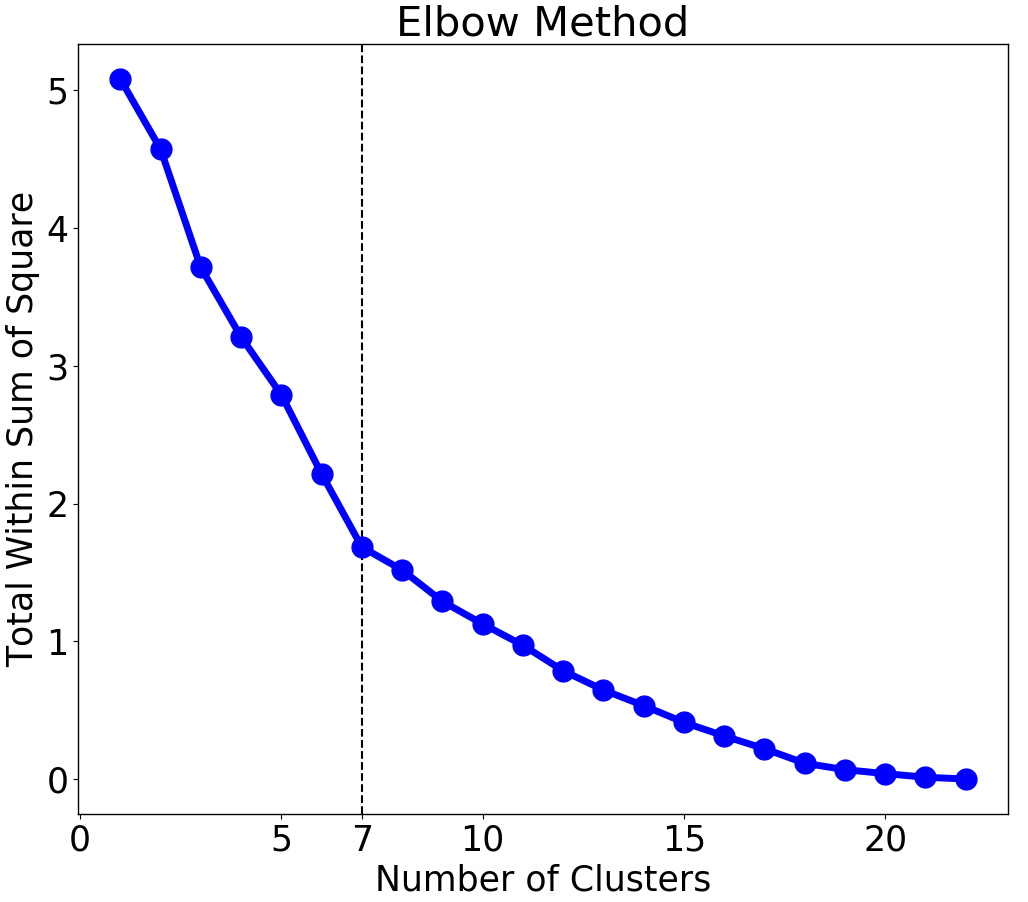}
  \caption{\small The optimal number of clusters determined by the elbow method for 23 languages$\to$English based on language embeddings. The elbow method plots the curve of clustering performance (which is defined as the intra-cluster variation, i.e., the within-cluster sum of squares) according to different number of clusters, and find the location of a bend (knee) in the curve as the optimal number of clusters (7 as shown in the figure).}
  \label{determin_cluster}
\end{figure}

\begin{itemize}
\item  Language embeddings capture the relationship in the language family well. Cluster \#7 in Figure~\ref{24_NMT_hirach_average_cosine_cluster5} roughly covers the Germanic and Romance languages (De, Nl, Ro, It, Fr, Es, Pt) which are two different language branches in the Indo-European family but both adopt Latin alphabets. It can be further exactly divided into Germanic (De, Nl) and Romance (Ro, It, Fr, Es, Pt) branches. 
Cluster \#6 is very close to Cluster \#7 as they all belong to the Indo-European family. The difference is that the languages in Cluster \#6 (Ru, El, Sk, Sl, Bg, Cs, Pl) mostly belong to Slavic branch\footnote{It also contains El (Greek) probably due to that most Slavic languages are influenced by Middle Greek through the Eastern Orthodox Church~\citep{horrocks2009greek}.}.

\item Language embeddings can also capture the knowledge of morphological typology~\citep{comrie1989language}. Ja (Japanese), Tr (Turkish), Hu (Hungarian) in Cluster \#4 of Figure~\ref{24_NMT_hirach_average_cosine_cluster5} are all Synthetic-Fusional languages. Language embedding based method can cluster them together by learning their language features end-to-end with the embeddings despite they are in different language families.


\item Language embeddings capture the regional, cultural, and historical influences. The languages in Cluster \#5 (He, Ar, Fa) of Figure~\ref{24_NMT_hirach_average_cosine_cluster5} are close to each other in geographical location (West Asia). Ar (Arabic) and He (Hebrew) share the same Semitic language branch in Afroasiatic language family, while Fa (Persian) has been influenced much by Arabic due to history and religion in West Asia\footnote{https://en.wikipedia.org/wiki/Persian\_vocabulary\#Arabic\_
influence}. Language embedding can implicitly learn this relationship and cluster them together.
\end{itemize}

\begin{table*}[t]
\small
\centering
\begin{tabular}{c c c c c c c c c c c c c}
\toprule
Language & Ar & Bg & Cs & De & El & Es & Fa & Fr & He & Hu &It &Ja  \\

\midrule
\textit{Random}  &22.90 & 32.18 &28.88 &30.67 &33.28 &28.47 &19.16 &24.36 &28.01 &20.78 &26.08 &10.13 \\
\textit{Family}  &25.02 &\textbf{32.75} &30.27 &31.09 &33.61 &28.18 &19.59 &24.24 &29.42 &19.07 &26.74 &9.90 \\
\midrule
\textit{Embedding}  &\textbf{25.27} &32.52 &\textbf{30.97} &\textbf{31.33} &\textbf{33.67} &\textbf{28.81} &\textbf{19.64} &\textbf{25.43} &\textbf{30.03} &\textbf{21.89} &\textbf{27.10} &\textbf{11.57} \\

\bottomrule
\bottomrule
Language &Nl &Pl &Pt &Ro &Ru & Sk & Sl & Th & Tr & Vi & Zh  \\
\midrule
\textit{Random} &33.88 &18.34 &31.93 &27.64 &17.38 &24.22 & 15.88 &17.94 &18.93 &25.93 & 14.08  \\
\textit{Family}  &34.82 &19.23 &32.18 &27.91 &17.58 &25.89 &\textbf{23.97} &\textbf{18.46} &19.95 &\textbf{26.95} &\textbf{15.13}  \\
\midrule
\textit{Embedding}  &\textbf{35.43} &\textbf{20.04} &\textbf{32.33} &\textbf{27.97} &\textbf{18.13} &\textbf{26.61} &22.12 &\textbf{18.46} &\textbf{22.09} &\textbf{26.95} &\textbf{15.13} \\

\bottomrule
\end{tabular}
\caption{\small BLEU score of 23 languages$\to$English with multilingual models based on different methods of language clustering: \textit{Random}, \textit{Family} (Language Family) and \textit{Embedding} (Language Embedding).} 
\label{bleu_score}
\end{table*}

These findings show that language embeddings can incorporate different prior knowledge such as language family, or some other implicit information like regional, cultural, and historical influence, and can well fuse the information together for clustering. In the next subsection, we show how our language embedding based clustering boosts the performance of multilingual NMT compared with the language family based clustering.

\begin{table*}[t]
\small
\centering
\begin{tabular}{c c c c c c c c c c c c c}
\toprule
Language  & Ar & Bg & Cs & De & El & Es & Fa & Fr & He & Hu &It &Ja  \\
\midrule
\textit{Data size} &180K &140K &110K &180K &180K &180K &70K &180K &150K &90K &180K &90K \\
\midrule

\textit{Individual} (23) &\textbf{25.43} &\textbf{32.87} &29.15 &\textbf{32.18} &\textbf{33.70} &\textbf{29.17} &18.12 &\textbf{27.98} &29.45 &19.07 &\textbf{27.70} &9.90 \\
\textit{Universal} (1) &23.26 &32.47 &28.86 &30.42 &33.56 &28.03 &19.45 &23.64 &27.29 &21.24 &26.07 &\textbf{12.78} \\

\textit{Embedding} (7)  &25.27 &32.52 &\textbf{30.97} &31.33 &33.67 &28.81 &\textbf{19.64} &25.43 &\textbf{30.03} &\textbf{21.89} &27.10 &11.57 \\
\bottomrule
\bottomrule
Language  &Nl &Pl &Pt &Ro &Ru & Sk & Sl & Th & Tr & Vi & Zh &  \\
\midrule
\textit{Data size} &140K &140K &180K &180K &160K &70K &14K &80K &130K &130K &180K \\
\midrule
\textit{Individual} (23) &34.61 &18.19 &\textbf{32.77} &27.88 &17.55 &19.72 &4.48 &\textbf{18.46} &19.95 &\textbf{26.95} &\textbf{15.13} \\
\textit{Universal} (1) &34.58 &19.02 &30.96 &27.77 &16.69 &25.31 &\textbf{24.22} &18.27 &18.76 &26.13 &14.54 \\
\textit{Embedding} (7)  &\textbf{35.43} &\textbf{20.04} &32.33 &\textbf{27.97} &\textbf{18.13} &\textbf{26.61} &22.12 &\textbf{18.46} &\textbf{22.09} &\textbf{26.95} &\textbf{15.13} \\
\bottomrule
\end{tabular}
\caption{\small BLEU score of 23 languages$\to$English with different number of clusters: \textit{Universal} (all the languages share one model), \textit{Individual} (each language with separate models, totally 23 models)), \textit{Embedding} (Language Embedding with 7 models). \textit{Data size} shows the training data for each language$\to$English.} 
\label{bleu_score_number}
\end{table*}

\subsection{Translation Accuracy}
\label{ret_multi_nmt}
After dividing the languages into several clusters, we train the models for the languages in each cluster with a multilingual NMT model~\citep{DBLP:journals/tacl/JohnsonSLKWCTVW17} based on different clustering methods and list the BLEU scores of each language$\to$English in Table~\ref{bleu_score}. We also show the results of random clustering (\textit{Random})  and use the same number of clusters as the language embedding based clustering, and average the BLEU scores on multiple times of random clustering (3 times in our experiments) for comparison. It can be seen that \textit{Random} performs worst, and language embedding (\textit{Embedding}) based clustering outperforms the language family (\textit{Family}) based clustering for most languages\footnote{\textit{Embedding} performs worse than \textit{Family} on only 2 languages.}, demonstrating the superiority of the learning based method over prior knowledge.

We further compare the language embedding based clustering with two extreme settings: 1) Each language trained with a separate model (\textit{Individual}), 2) All languages trained in a universal model (\textit{Universal}), to analyze how language clustering performs on each language. As shown in Table~\ref{bleu_score_number}, we have several observations:

\begin{itemize}
\item First, \textit{Universal} in general performs the worst  due to the diverse languages and limited model capacity. \textit{Universal} only performs better on Ja (Japanese) and Sl (slovene) due to their small amount of training data, which will be discussed in the next observation.

\item Second, \textit{Embedding} outperforms \textit{Individual} (mostly by more than 1 BLEU score) on 12 out of 23 languages\footnote{\textit{Embedding} performs slightly worse (mostly within 0.5 BLEU gap) than \textit{Individual} on 8 languages, mainly due to the plenty of training data on these languages. In this case, \textit{Individual} with more number of models will help.}, which demonstrates that clustering can indeed bring performance gains with fewer models, especially for languages with relatively small data size. For these languages, clustering together with similar languages may act as some kind of data augmentation, which will benefit the model training. An extreme case is that Sl (slovene) training together with all the languages gets the best accuracy as shown in Table~\ref{bleu_score_number}, since Sl is extremely low-resource, which benefits a lot from more training data. This can be also verified by the next observation.

\item Third, similar languages can help boost performance. For example, Hu (Hungarian) and Ja (Japanese) are similar as they are typical agglutinative languages in terms of morphological typology~\citep{comrie1989language}. When they are clustered together based on \textit{Embedding}, the performance of both languages improves (Hu: 19.07$\to$21.89, Ja: 9.90$\to$11.57). 

\end{itemize}

\begin{figure}[!b]
    \centering
    \begin{subfigure}{0.47\textwidth} 
        \includegraphics[width=\textwidth]{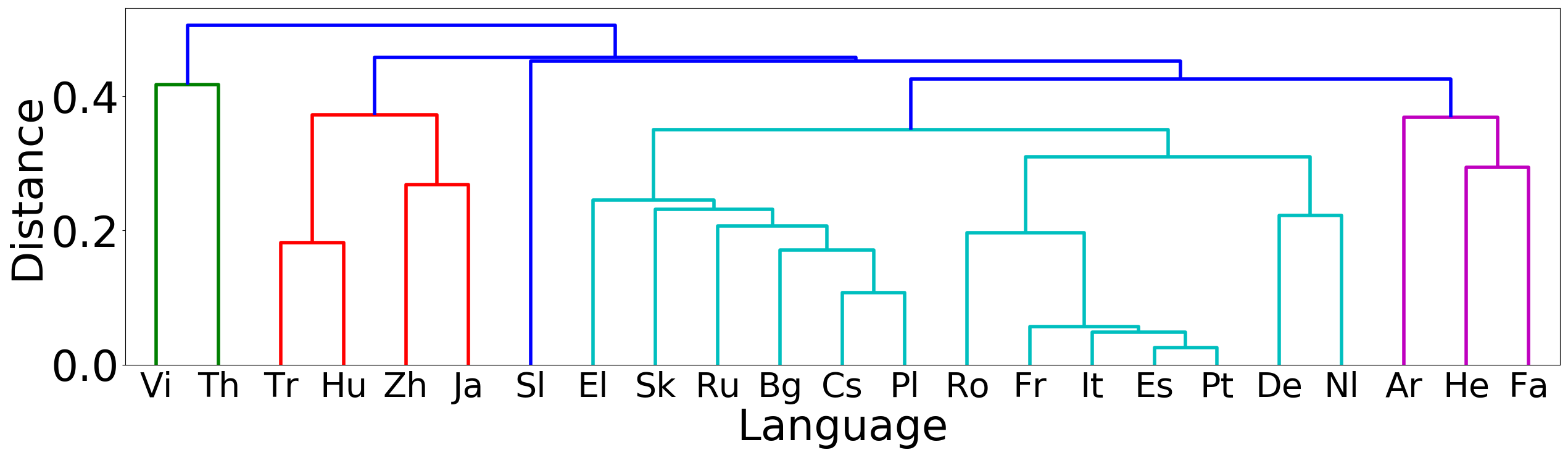}
        \caption{50\% of training data} 
        \label{cluster_varying_data_50}
    \end{subfigure}
    \vspace{0.2em} 
    \begin{subfigure}{0.47\textwidth} 
        \includegraphics[width=\textwidth]{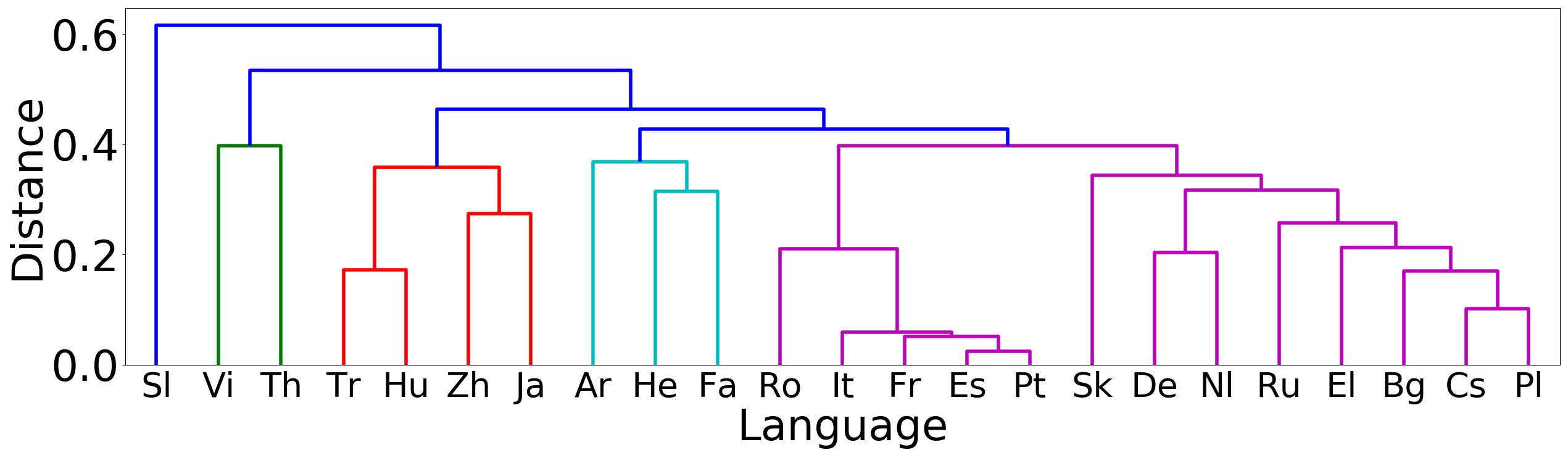}
        \caption{20\% of training data} 
        \label{cluster_varying_data_20}
    \end{subfigure}
      \begin{subfigure}{0.47\textwidth} 
        \includegraphics[width=\textwidth]{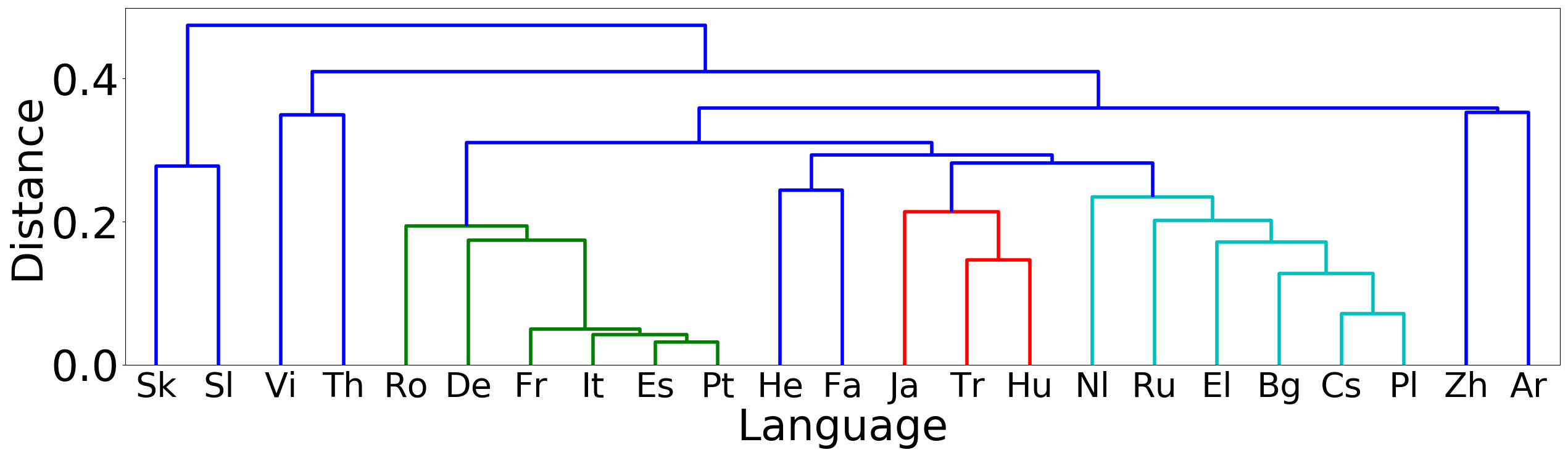}
        \caption{5\% of training data} 
        \label{cluster_varying_data_5}
    \end{subfigure}  
     \caption{\small The results of language embedding based clustering with varying amount of training data.} 
   \label{cluster_varying_data}
\end{figure}

\subsection{Clustering Results with respect to Training Data Scale}
\label{ret_vary_data}
We further study how the clustering results based on language embeddings change with varying training data, in order to study the robustness of our language embedding based clustering, which is important in the low-resource multilingual setting. We reduce the training data of each language to 50\%, 20\% and 5\% to check how stable the language embeddings based clustering is, as shown in Figure~\ref{cluster_varying_data}. It can be seen that most languages are clustered stably when training data is relatively large (50\% or even 20\%), except for Sl (Slovene) which has the least training data (14K) among the 23 languages and thus reducing the training data of Sl to 50\% and 20\% will influence the learning of language embeddings and further influence clustering results. If further reducing the training data to 5\%, more languages are influenced and clustered abnormally as most languages have less than 10K training data. Even so, the similar languages such as Fr (French), It (Italian), Es (Spanish), and Pt (Portuguese) are still clustered together.

\begin{figure*}
  \centering
  \includegraphics[width=1.0\textwidth]{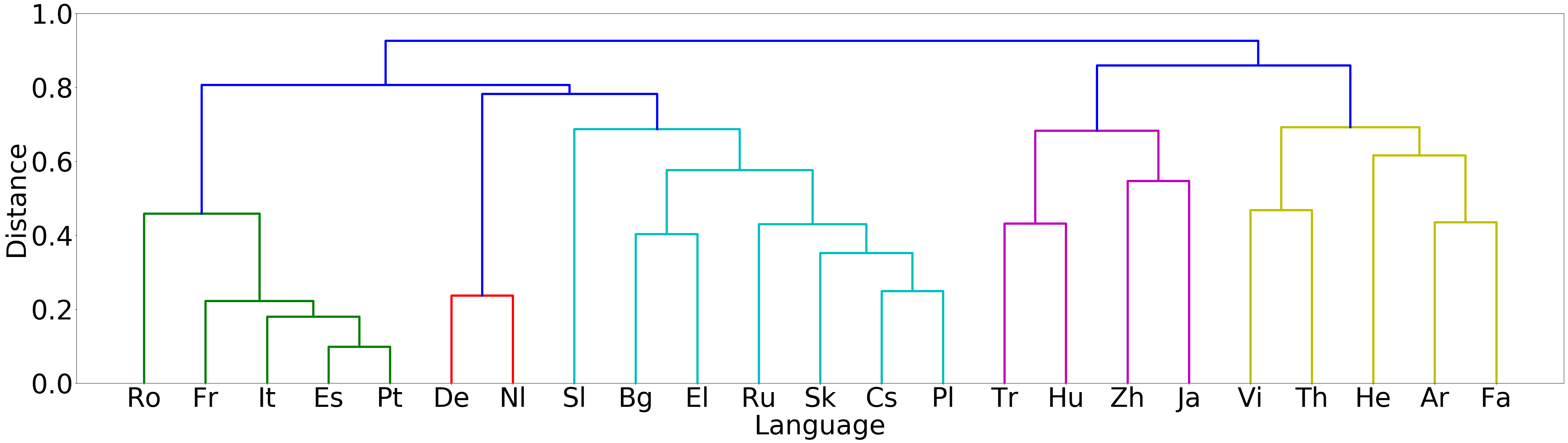}
  \caption{The hierarchical clustering based on language embedding in one-to-many setting. The Y-axis represents the distance between two languages or clusters. Languages in the same color are divided into the same cluster. Cluster \#1: Ro, Fr, It, Es, Pt; Cluster \#2: De, Nl; Cluster \#3: Sl, Bg, El, Ru, Sk, Cs, Pl; Cluster \#4: Tr, Hu, Zh, Ja; Cluster \#5:  Vi, Th, He, Ar, Fa.}
  \label{24_hirach_average_cosine_cluster_enxx}
\end{figure*}

\subsection{Results of One-to-Many Translation}
\label{ret_one_to_many}

\begin{table*}[!t]
\small
\centering
\begin{tabular}{c c c c c c c c c c c c c}
\toprule
Language & Ar & Bg & Cs & De & El & Es & Fa & Fr & He & Hu &It &Ja  \\
\midrule
\textit{Universal} &9.80 &25.84 &17.89 &22.40 &26.71 &28.48 &12.26 &22.10 &16.38 &13.32 &25.34 &10.91 \\
\textit{Individual} &\textbf{13.13} &\textbf{30.04} &19.84 &\textbf{25.69} &\textbf{27.90} &\textbf{29.57} &12.03 &\textbf{22.93} &\textbf{20.43} &13.74 &\textbf{26.87} &10.70 \\
\midrule
\textit{Family}  &13.11 &27.54 &19.11 &23.74 &27.78 &28.88 &13.36 &22.78 &20.26 &13.74 &26.71 &10.70 \\
\textit{Embedding}  &12.37 &28.85 &\textbf{20.81} &25.27 &27.11 &28.93 &\textbf{13.79} &22.85 &19.23 &\textbf{15.47} &26.81 &\textbf{13.33} \\
\bottomrule
\vspace{0.1em} \\
\bottomrule
Language &Nl &Pl &Pt &Ro &Ru & Sk & Sl & Th & Tr & Vi & Zh &  \\
\midrule
\textit{Universal} &26.19 &10.10 &28.27 &19.72 &8.26 &15.52 &\textbf{15.82} &25.06 &9.44 &26.87 &9.56 \\
\textit{Individual} &29.86 &11.61 &28.09 &\textbf{21.81} &\textbf{14.11} &14.47 &6.61 &27.41 &11.40 &28.77 &\textbf{10.83} \\
\midrule
\textit{Family}  &27.83 &10.97 &28.63 &21.13 &12.80 &16.91 &15.73 &27.41 &11.40 &28.77 &\textbf{10.83} \\
\textit{Embedding}  &\textbf{29.98} &\textbf{11.95} &\textbf{28.83} &21.14 &13.84 &\textbf{18.18} &14.25 &\textbf{28.55} &\textbf{12.11} &\textbf{29.79} &10.52 \\
\bottomrule
\end{tabular}
\caption{BLEU score of English$\to$23 languages with multilingual models based on different methods of language clustering: \textit{Universal} (all the languages share one model), \textit{Individual} (each language with separate model), \textit{Family} (Language Family), \textit{Embedding} (Language Embedding).} 
\label{bleu_score_one_to_many}
\end{table*}

We finally show experiment results of one-to-many translation, i.e., English$\to$23 languages. We first train a universal model that covers all languages to get the language embeddings, and then perform clustering like in the many-to-one setting. The optimal number of clusters is 5, which is automatically determined by the elbow method (we show the figure of the elbow method in Figure~1 in the Supplementary Materials).  

Figure~\ref{24_hirach_average_cosine_cluster_enxx}  shows the clustering results based on language embeddings, and Table~\ref{bleu_score_one_to_many} shows the BLEU score of multilingual models clustered by different methods. We also have several findings:

\begin{itemize}
\item  In one-to-many setting, language embeddings still depict the relationship among language families. Cluster \#1 in Figure~\ref{24_hirach_average_cosine_cluster_enxx} covers the Romance (Ro, Fr, It, Es, Pt) branch in Indo-European language family while Cluster \#2 contains the Germanic (De, Nl) branch in Indo-European language family. The languages in Cluster \#3 (Sl, Bg, El, Ru, Sk, Cs, Pl) are mostly from the Slavic branch in Indo-European language family. We find that the cluster results of Indo-European language family is more fine-grained than many-to-one setting, which divides each language into their own branches in a language family. 

\item Language embedding can capture regional, cultural, and historic influence. For Cluster \#4 in Figure~\ref{24_hirach_average_cosine_cluster_enxx}, Zh (Chinese) and Ja (Japanese) are in different language families but influence each other through the history and culture. Th (Thai) and Vi (Vietnamese) in Cluster \#5 that are close to each other in geographical location can be captured by their language embeddings.

\end{itemize}

These findings show that language embeddings can stably characterize relationships among languages no matter in one-to-many setting or in many-to-one setting. Although there exist some differences in the clustering results between one-to-many setting and many-to-one setting due to the nature of the learning based clustering method, most of the clustering results are reasonable. Our experiment results shown in Table~\ref{bleu_score_one_to_many} demonstrate that the language embedding can provide more reasonable clustering than prior knowledge, and thus result in higher BLEU score for most of the languages. 

We also find from Table~\ref{bleu_score_one_to_many} that each language with separate model (\textit{Individual}) perform best on nearly 10 languages, due to the abundant model capacity. However, there are still several languages on which our clustering methods (\textit{Family} or \textit{Embedding}) largely outperform \textit{Individual}. The reason is similar as the many-to-one setting that some languages are of smaller data sizes. In this situation, similar languages clustered together will boost the accuracy compared with separate model.

\section{Conclusion}
In this work, we have studied language clustering for multilingual neural machine translation. Experiments on 23 languages$\to$English and English$\to$23 languages show that language embeddings can sufficiently characterize the similarity between languages and outperform prior knowledge (language family) for language clustering in terms of the BLEU scores. 

\begin{table*}[t]
\small
\centering
\begin{tabular}{c|c c c c c c c c c c c c c c c}
\toprule
Language & Ar & Bg & Cs & De & El & Es & Fa & Fr & He & Hu &It &Ja \\
\midrule
Data size  &180K &140K &110K &180K &180K &180K &70K &180K &150K &90K &180K &90K \\
\midrule
\midrule
Language &Nl &Pl &Pt &Ro &Ru & Sk & Sl & Th & Tr & Vi & Zh &  \\
\midrule
Data size  &140K &140K &180K &180K &160K &70K &14K &80K &130K &130K &180K \\
\bottomrule
\end{tabular}
\caption{The size of training data for 23 language$\leftrightarrow$English in our experiments.}
\label{data_size}
\end{table*}

\begin{table*}[t]
\small
\centering
\begin{tabular}{c|c c c c c c c c c }
\toprule
Language & Arabic &Bulgarian &Czech &German &Greek &Spanish &Persian &French &Hebrew \\
\midrule
ISO code  &Ar &Bg &Cs &De &El &Es &Pt &Fr &He \\
\midrule
\midrule
\textit{Language} &Hungarian &Italian &Japanese &Dutch &Polish &Portuguese &Romanian &Russian &Slovak  \\
\midrule
ISO code  &Hu &It &Ja &Nl &Pl &Pt &Ro &Ru &Sk \\
\midrule
\midrule
\textit{Language} &Slovenian &Thai &Turkish &Vietnamese &Chinese & English \\
\midrule
ISO code &Sl &Th &Tr &Vi &Zh  & En\\
\bottomrule
\end{tabular}
\caption{The ISO 639-1 code of each language in our experiments.}
\label{language_code}
\end{table*}

For future work, we will test our methods for many-to-many translation. We will consider more languages (hundreds or thousands) to study our methods in larger scale setting. We will also study how to obtain language embeddings from monolingual data, which will make our method scalable to cover those languages with little or without bilingual data. On the other hand, we will also consider other pre-training methods~\citep{song2019mass} for multilingual and low-resource NMT.

\appendix
\section{Dataset Description}
We evaluate our experiments on IWSLT datasets. We collect 23 languages$\leftrightarrow$English translation pairs form IWSLT evaluation campaign from year 2011 to 2018. The training data sizes of each languages are shown in Table~\ref{data_size} and we use the default validation and test set for each language pairs. For Bg (Bulgarian), El (Greek), Hu (Hungarian) and Ja (Japanese), there are no available validation and test data in IWSLT, we randomly split 1K sentence pairs from the corresponding training set as the validation and test data respectively. 

\section{Language Name and Code}
The language names and their corresponding language codes according to ISO 639-1 standard are listed in Table~\ref{language_code}.

\section{The Elbow Method for One-to-Many Setting}
The optimal number of clusters is 5 in one-to-many setting, which is automatically determined by the elbow method as shown in Figure~\ref{determin_cluster}.
\begin{figure}[htbp]
  \centering
  \includegraphics[width=0.4\textwidth]{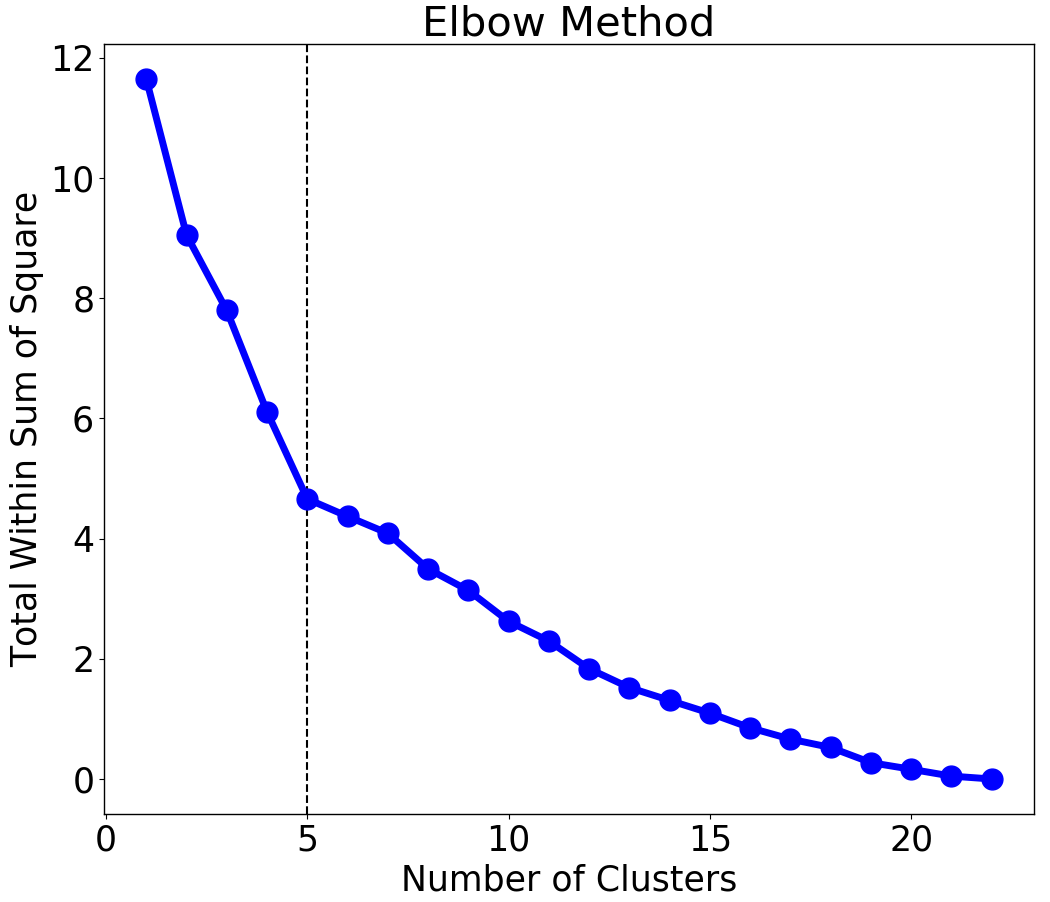}
  \caption{The optimal number of clusters determined by the elbow method for English$\to$23 languages based on language embeddings. The detailed description of the elbow method can be seen in Figure 4 in the main paper.}
  \label{determin_cluster}
\end{figure}

\bibliography{emnlp-ijcnlp-2019}
\bibliographystyle{acl_natbib}

\end{document}